\documentclass[10pt,twocolumn,letterpaper]{article}

\usepackage[pagenumbers]{cvpr} 

\usepackage[dvipsnames]{xcolor}

\definecolor{cvprblue}{rgb}{0.21,0.49,0.74}
\usepackage[pagebackref,breaklinks,colorlinks,citecolor=cvprblue]{hyperref}

\usepackage{makecell}
\usepackage{graphicx}
\usepackage{amsmath, bm}
\usepackage{tabularx,booktabs}
\usepackage{adjustbox}
\usepackage{multirow}
\usepackage{graphicx,xcolor,colortbl} 
\definecolor{LightCyan}{rgb}{0.88,1,1}
\usepackage{array}
\newcolumntype{C}[1]{>{\centering\arraybackslash}p{#1}}
\newcolumntype{L}[1]{>{\arraybackslash}p{#1}}
\usepackage{subcaption}
\usepackage{indentfirst}
\usepackage{pifont}

\title{PatchFusion: An End-to-End Tile-Based Framework \\
for High-Resolution Monocular Metric Depth Estimation}

\author{%
Zhenyu Li, Shariq Farooq Bhat, Peter Wonka \\
King Abdullah University of Science and Technology (KAUST) \\
\small\url{https://zhyever.github.io/patchfusion/} \\
{\tt\small zhenyu.li.1@kaust.edu.sa}
}

\begin{document}

\twocolumn[{%
\renewcommand\twocolumn[1][]{#1}%
\maketitle
\begin{center}
    \centering
    \captionsetup{type=figure}
    \includegraphics[width=\textwidth]{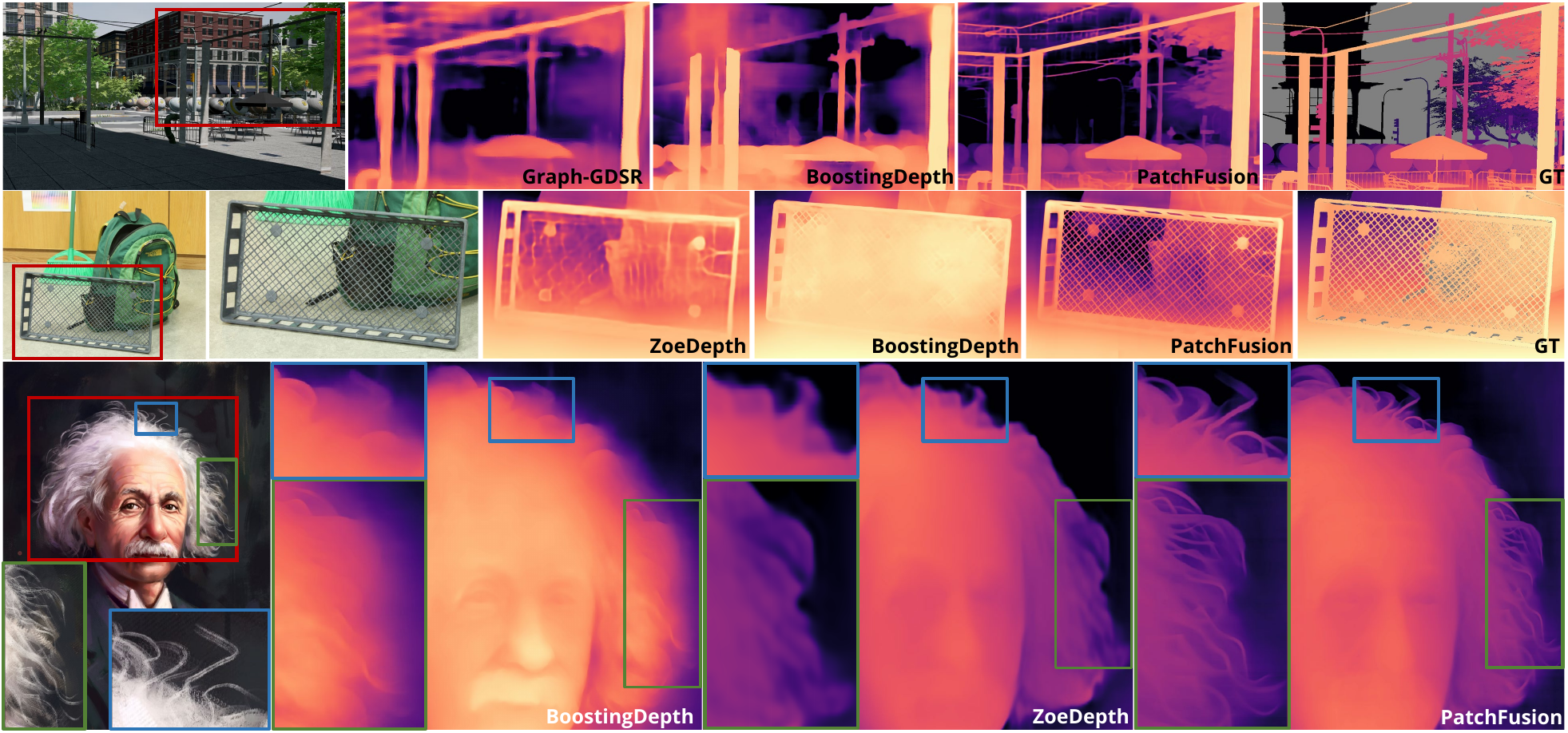}
    \captionof{figure}{\textbf{High-Resolution Depth Estimation.} Our tile-based monocular metric depth estimation model processes high-resolution images and delivers high-quality depth estimations with intricate details on test images corresponding to the synthetic training dataset UnrealStereo4K~\cite{tosi2021smd} as well as for zero-shot generalization to other types of real images. \textbf{Top:}~In-domain sample from UnrealStereo4K. \textbf{Middle:} Out-of-domain sample from Middleburry 2014~\cite{scharstein2014mid}. \textbf{Bottom:} Out-of-domain sample from the internet.}
    \label{fig:teaser}
\end{center}%
}]

\begin{abstract}

Single image depth estimation is a foundational task in computer vision and generative modeling. However, prevailing depth estimation models grapple with accommodating the increasing resolutions commonplace in today's consumer cameras and devices. 
Existing high-resolution strategies show promise, but they often face limitations, ranging from error propagation to the loss of high-frequency details. We present \textbf{PatchFusion}, a novel tile-based framework with three key components to improve the current state of the art: \textbf{(1)} A patch-wise fusion network that fuses a globally-consistent coarse prediction with finer, inconsistent tiled predictions via high-level feature guidance, \textbf{(2)} A Global-to-Local (G2L) module that adds vital context to the fusion network, discarding the need for patch selection heuristics, and \textbf{(3)} A Consistency-Aware Training (CAT) and Inference (CAI) approach, emphasizing patch overlap consistency and thereby eradicating the necessity for post-processing. 
Experiments on UnrealStereo4K, MVS-Synth, and Middleburry 2014 demonstrate that our framework can generate high-resolution depth maps with intricate details. PatchFusion is independent of the base model for depth estimation. Notably, our framework built on top of SOTA ZoeDepth brings improvements for a total of 17.3\% and 29.4\% in terms of the root mean squared error (RMSE) on UnrealStereo4K and MVS-Synth, respectively.

\end{abstract}

\section{Introduction}
\label{sec:intro}

This paper addresses the challenge of metric single image depth estimation for high-resolution inputs. Single image depth estimation remains a cornerstone in computer vision and generative modeling~\cite{eigen2014mde,bhat2023zoedepth,li2022binsformer,zhang2023controlnet}. Yet, most state-of-the-art (SOTA) depth estimation architectures are bottlenecked by the resolution capabilities of their backbone. For instance, ZoeDepth~\cite{bhat2023zoedepth} processes an input resolution of 384$\times$512, VPD~\cite{zhao2023unleashing} manages 480$\times$480, and AiT~\cite{ning2023all} is designed for 384$\times$512. These figures pale in comparison to the resolutions offered by modern consumer cameras, such as the 45 Megapixel Canon EOS R5, the widely available 8K televisions, and even mobile devices like the iPhone 15, which boasts a 12MP Ultra Wide lens.

Several methods have attempted to bridge this resolution gap:
\textbf{(1) Guided Depth Super-Resolution (GDSR)} techniques~\cite{zhao2022dgsrdiscrete,metzger2023gdsrdiff,hui2016gdsrdepth,zhong2023guided} aim to refine high-resolution depth maps from their low-resolution counterparts using high-resolution color images as reference.
\textbf{(2) Implicit Function} approaches, such as SMD-Net~\cite{tosi2021smd}, leverage algorithms like~\cite{mildenhall2021nerf,chen2021liif} to estimate disparities continuously across image locations, essentially performing on-the-fly super-resolution. However, due to the low-resolution nature of many models, these techniques have their limitations: GDSR can propagate errors, and the implicit function still strips away crucial high-frequency details during input downsampling.
Lastly, the \textbf{(3) Tile-Based Method} proposed in BoostingDepth~\cite{miangoleh2021boostingdepth}, emphasizes \textit{relative} depth estimation by processing image patches independently before merging them to form a unified depth map.

Our approach refines the concept of tile-based depth estimation. While the BoostingDepth already has promising results, we identified some shortcomings to improve. First, we discover that BoostingDepth suffers from scale inconsistencies, especially when transposed to \textit{metric} depth estimation. These inconsistencies then mandate rigorous post-process corrections, such as scale optimization and Gaussian blending. Second, the fusion network in BoostingDepth often stumbles due to the lack of guidance and its inability to grasp a more holistic view of the input image, leading to local optima and compelling the use of complex heuristic patch selections.

In light of these challenges, we introduce \textbf{PatchFusion} that stands on three pillars:
\textbf{Firstly}, we augment the fusion network with high-level feature guidance, streamlining its training. \textbf{Secondly}, our proposed Global-to-Local (G2L) module empowers the fusion network to stay context-aware, eliminating complex patch selection heuristics. \textbf{Thirdly}, our Consistency-Aware Training (CAT) and Inference (CAI) strategy places a special emphasis on patch overlap consistency, facilitating consistency-aware training and inference. As a result, we achieve an end-to-end tile-based framework without any necessity for pre-processing such as patch selection, and post-processing like scale optimization and Gaussian blending. In Fig.~\ref{fig:teaser} we show selected examples to illustrate how PatchFusion is able to significantly improve the results compared to the previous SOTA. In summary, our key contributions include:

\begin{itemize}
\item The introduction of a novel tile-based network architecture and training strategy for metric monocular depth estimation called PatchFusion. PatchFusion is adept at handling high-resolution images and is the first tile-based metric depth estimations approach that can be trained in an end-to-end manner without the need for additional post-processing or pre-processing steps.
\item We conducted exhaustive empirical validations using the UnrealStereo4K~\cite{tosi2021smd}, MVS-Synth~\cite{huangDeepMVS2018}, and Middleburry 2014~\cite{scharstein2014mid} datasets. Our framework further improves current SOTA by 17.3\% and 29.4\% on UnrealStereo4K and MVS-Synth, respectively.
\end{itemize}

\section{Related Work}

\subsection{Monocular Depth Estimation}
Tremendous progress in monocular depth estimation has been achieved by publicly available large-scale datasets~\cite{silberman2012nyu,geiger2013kitti,dai2017scannet,cordts2016cityscapes}, network design~\cite{eigen2014mde,bhat2021adabins,li2022binsformer,wang2020cliffnet,yang2021transformer,li2023depthformer}, loss supervision~\cite{lee2020multiloss,liu2023va,xian2020rankloss}, refined problem formulations~\cite{fu2018dorn,diaz2019soft,bhat2021adabins,xian2020rankloss}, and training strategies~\cite{petrovai2022pseudomde,godard2019mde2,fan2023contrastive}.
Recently, the best-performing frameworks have been built on the transformer architecture~\cite{liu2021swin,bao2021beit,dosovitskiy2020vit}.
While current SOTA frameworks demonstrate exceptional performance, they still use low-resolution images as input.
For example, the SOTA ZoeDepth~\cite{bhat2023zoedepth} uses BEIT$_{384}$-L~\cite{bao2021beit} with 307M parameters and only infers 384$\times$512 (about 0.2 megapixel) images.
This stands in stark contrast to the advancements in modern imaging devices that capture images at increasing resolutions, and the growing demand among users for high-resolution depth estimation. In this work, we aim to utilize these large-scale models on high-resolution inputs and achieve high-resolution depth estimation.

\begin{figure*}[t]
    \centering
    \begin{subfigure}{.35\textwidth}
        \centering
        \includegraphics[width=1\linewidth]{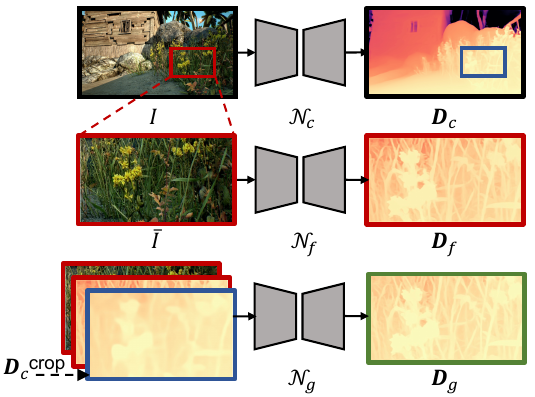}
        \caption{Overall Pipeline}
        \label{fig:figsub:pipeline}
    \end{subfigure}%
    \begin{subfigure}{.64\textwidth}
        \centering
        \includegraphics[width=.9\linewidth]{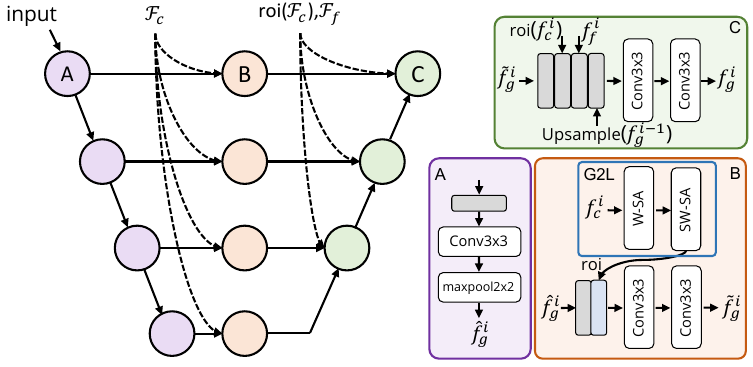}
        \caption{Guided Fusion Network}
        \label{fig:figsub:gfn}
    \end{subfigure}
    \caption{\textbf{Framework Illustration.} \textbf{(a) Overall Pipeline:} The framework consists of three phases, including a coarse network $\mathcal{N}_{c}$ providing globally consistent yet coarse depth estimation $\mathbf{D}_{c}$, a fine network $\mathcal{N}_{f}$ presenting finer details for an input tile (but all tiles together would lack consistency), and a guided fusion network $\mathcal{N}_{g}$ combining the best of both worlds. \textbf{(b) Architecture of Guided Fusion Network:} The lightweight network includes (A) successive encoder layers, (B) skip-connection modules, and (C) upsampling layers.}
    \label{fig:overall}
\end{figure*}

\subsection{High-Resolution Depth}
The pursuit of high-resolution depth estimation has typically converged on three prevalent strategies:
\textbf{(1) Guided Depth Super-Resolution (GDSR)}: This approach employs methods~\cite{zhao2022dgsrdiscrete,metzger2023gdsrdiff,hui2016gdsrdepth,zhong2023guided} to reconstruct high-resolution depth maps using low-resolution depth observations. These reconstructions are facilitated by paired high-resolution color images. A notable limitation of GDSR is its reliance on training samples derived from the direct downsampling of high-resolution depth maps. When applied to monocular depth estimation—where low-resolution depth maps stem from models limited by input resolution constraints—this approach can introduce cascading errors during the super-resolution process. \textbf{(2) Implicit Function}: Melding the implicit function~\cite{mildenhall2021nerf,chen2021liif} with stereo mixture density has enabled architectures like SMD-Net~\cite{tosi2021smd} to deliver precise disparity estimations across continuous image locations, thereby addressing the stereo super-resolution challenge. Nevertheless, this method hasn't fully confronted the constraints posed by image resolution. Models are still vulnerable to losing intricate high-frequency details during the input downsampling phase. \textbf{(3) Tile-Based Method}: This paper aligns itself with the tile-based strategy, pioneered in works such as BoostingDepth~\cite{miangoleh2021boostingdepth} and \cite{rey2022360monodepthtile}. The underlying principle here is to segment depth estimation into patches. Subsequently, these patches are merged to construct a holistic image depth estimation. A chief advantage of this method is its ability to sidestep the severe downsampling often mandated by the restrictions on input resolution.

\subsection{Depth Map Blending}
Given coarse whole-image and fine patch-wise depth maps, blending them together is a crucial part of tile-based methods.
This process aims to keep the correct global scale in coarse maps while maintaining the fine details in the tile maps.
BoostingDepth~\cite{miangoleh2021boostingdepth} applies a linear polynomial, whereas \cite{rey2022360monodepthtile} utilizes a deformable depth field proposed in \cite{hedman2018instant} to achieve the blending.
The former optimizes one patch-wise scale and shift in the least squares manner and the latter calculates pixel-wise deformation. Unlike previous approaches, the output of our method can be seamlessly stitched without the necessity for any post-optimization, resulting in an end-to-end framework for high-resolution depth estimation.

\section{Method}
\label{sec:method}

In this section, we present the overall framework in Sec.~\ref{subsec:framework}, Consistency-Aware Training and Inference in Sec~\ref{subsec:cat}, and implementation details in Sec.~\ref{subsec:train&infer}.

\subsection{Overall Framework}
\label{subsec:framework}
Our primary objective is to harness the capabilities of a pre-trained base depth estimation model trained on low resolution and employ it for high-resolution depth estimation, targeting high resolutions, e.g. 4K. 
Unfortunately, a straightforward scaling of current base models to high resolutions such as 4K far exceeds the memory and compute capabilities of current hardware.
Therefore, we adopt a patch-wise approach, breaking down the high-resolution depth estimation task into three distinct steps (see Fig.~\ref{fig:figsub:pipeline}): \textbf{(i)} Global Scale-Aware Estimation, \textbf{(ii)} Local Fine-Depth Estimation, and \textbf{(iii)} Fusion. We train a dedicated network for each step as described below. 

\textbf{(i) Global Scale-Aware Estimation:} In the initial step, we predict a coarse depth map by first downsampling the input image to the native resolution of the depth model. This downsampling reduces the memory requirements and computational load while providing an initial estimation of the depth, $\mathbf{D}_{c}$. For this step, we fine-tune our base depth model on the downsampled data, resulting in the coarse network $\mathcal{N}_{c}$. Due to downsampling, the output from this step is coarse in nature and fine, high-frequency details are lost at the cost of global consistency.

\begin{figure}
    \centering
    \includegraphics[width=1.0\linewidth]{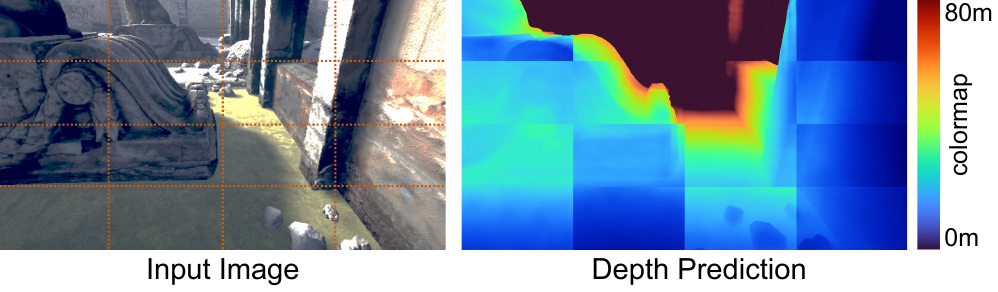}
    \caption{\textbf{Illustration of Inconsistent Depth Prediction.} Patch-wise depth prediction by itself suffers from the lack of global information, leading to inconsistent depth predictions especially visible at the patch boundaries.}
    \label{fig:inconsistency}
\end{figure}

\textbf{(ii) Patch-Wise Depth Prediction:} In this step, we divide the input images into smaller manageable patches and feed the cropped patches as input to our base model. We use a fixed patch size that is equal to or similar to the native resolution of the base depth model. This results in a fine prediction $\mathbf{D}_{f}$ containing rich details, particularly at boundaries and intricate structures.
Nevertheless, this detailed depth map, being confined to only a segment of the original scene, remains oblivious to the global context, making its scale potentially inconsistent with the actual scene. This could result in a scale-shifted prediction and fluctuations across patches leading to obvious patch artifacts as shown in Fig.~\ref{fig:inconsistency}.

\textbf{(iii) Fusion and Guided Fusion Network}:
To resolve the inconsistency issues with patch-wise prediction, our goal in this step is to transfer the global information from the coarse depth obtained in \textbf{(i)} to the patch-wise predictions without compromising their higher details. While the transfer could be implemented by simply learning a pix2pix U-Net~\cite{ronneberger2015unet, miangoleh2021boostingdepth}, our key idea is to exploit the multi-scale features from $\mathcal{N}_{c}$ and $\mathcal{N}_{f}$. We use two main components to achieve this transfer - the Global-to-Local Module (G2L) and the Guided Fusion Network $\mathcal{N}_{g}$.

\textbf{Global-to-Local (G2L) Module}: Our empirical evaluations (in Tab.~\ref{tab:abl}) indicate that directly adopting the global guidance feature $\mathcal{F}_c=\{f_c^i\}_{i=1}^{L}$ from $\mathcal{N}_{c}$ (where $L$ is the number of layers) still suffers from the scale-shift issue. Even though $\mathcal{F}_c$ is derived from the entirety of the image, the necessary information needed for accurate scale inference during fusion is lost post the cropping operation. Addressing this, we present our G2L module designed to retain global context.

While the key insight of G2L is to apply the global-wise self attention for each-level feature in $\mathcal{F}_c$ to ensemble crucial information for patch-wise scale-consistent prediction, we adopt the Swin Transformer Layer (STL)~\cite{liu2021swin} to preserve the global context while simultaneously alleviating GPU memory concerns. 
The main ideas are the local attention and the shifted window mechanism. Given each feature map $f_c^i$, STL subdivides it into localized windows for self-attention (W-SA), which is then followed by shifted-window attention for inter-window interactions (SW-SA). The operation can be formulated as:
\begin{equation}
\label{eq:g2l}
    f_{g2l}^i = \texttt{G2L}(f_c^i) = \texttt{SW-SA}(\texttt{W-SA}(f_c^i)),
\end{equation}
where the superscript $i$ iterates through the multi-layer features. The output set $\mathcal{F}_{g2l}=\{f_{g2l}^i\}_{i=1}^{L}$ is then sent to the Guided Fusion Network $\mathcal{N}_{g}$ for further fusion.

\textbf{Guided Fusion Network}: The guided fusion network follows the U-Net~\cite{ronneberger2015unet} design as shown in Fig.~\ref{fig:figsub:gfn}. The input comprises a concatenated ensemble of the cropped original image $I$, the corresponding cropped coarse depth estimations $\mathbf{D}_{c}$ from $\mathcal{N}_{c}$, and fine depth estimations $\mathbf{D}_{f}$ from $\mathcal{N}_{f}$. The key idea of the design is to only use image and depth values in the encoder and to delay the injection of network features to the skip connections and decoder layers.

We use a lightweight encoder consisting of successive convolutional and max-pooling layers to extract multi-level features.
During the skip-connection, we inject the scale-aware feature $\mathcal{F}_{g2l}$ with a fusion block ($\texttt{FB}$) consisting of two $3\times 3$ convolutional layers with $\texttt{ReLU}$ activation functions as
\begin{equation}
\label{eq:g2l_fuse}
    \Tilde{f}_g^i = \texttt{FB}(\texttt{roi}(f_{g2l}^i), \hat{f}_g^i),
\end{equation}
where we apply the $\texttt{roi}$~\cite{he2017mask} operation to fetch features of the corresponding cropped area. $\hat{f}_g^i$ denotes the initial output feature from the $i$-th encoder layer.
As for the decoder, we again harness the guidance features $\mathcal{F}_c=\{f_c^i\}_{i=1}^{L}$ and $\mathcal{F}_f=\{f_f^i\}_{i=1}^{L}$ from $\mathcal{N}_{c}$ and $\mathcal{N}_{f}$, respectively, integrating them into the decoder of our fusion model. The operation can be formulated as:
\begin{equation}
\label{eq:fusion}
    f_g^i = \texttt{FB}(\Tilde{f}_g^{i}, \texttt{roi}(f_c^i), f_f^i, \texttt{Upsample}(f_g^{i-1}))
\end{equation}
where $\Tilde{f}_g^i$ is obtained from Eq.~\ref{eq:g2l_fuse}. The $\texttt{Upsample}$ function 2$\times$ rescales the features from the previous level in the decoder. The output features $\mathcal{F}_g=\{f_g^i\}_{i=1}^{L}$ are then sent to a depth head~\cite{bhat2023zoedepth} for final depth estimation.

\begin{figure}
    \centering
    \includegraphics[width=0.7\linewidth]{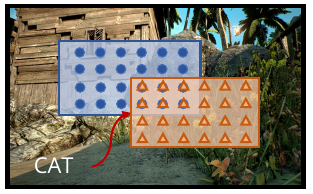}
    \caption{\textbf{Consistency-Aware Training Illustration.} We apply our consistency loss on the overlap area of intermediate features and depth predictions from two patches.}
    \label{fig:cl}
\end{figure}

\subsection{Consistency-Aware Training and Inference}
\label{subsec:cat}

The effectiveness of the fusion network hinges on not just the accuracy of predictions, but also their consistency across patch boundaries. While our Guided Fusion Network makes scale-aware predictions, boundary inconsistencies still exist. Recognizing this gap, we introduce Consistency-Aware Training (CAT) and Inference (CAI) to ensure patch-wise depth prediction consistency.

\textbf{Training}:
Our methodology, see Fig.~\ref{fig:cl}, is based on the intuitive idea that overlapping regions between cropped patches from the same image should ideally produce consistent feature representations and depth predictions. We begin by cropping an image patch, denoted as $I_1$. By shifting the cropping window, we obtain another cropped patch $I_2$, such that there exists an overlap region $\Omega$.
\begin{figure*}[h]
\setlength\tabcolsep{1pt}
\centering
\small
    \begin{tabular}{@{}*{5}{C{3.4cm}}@{}}
    \includegraphics[width=1\linewidth]{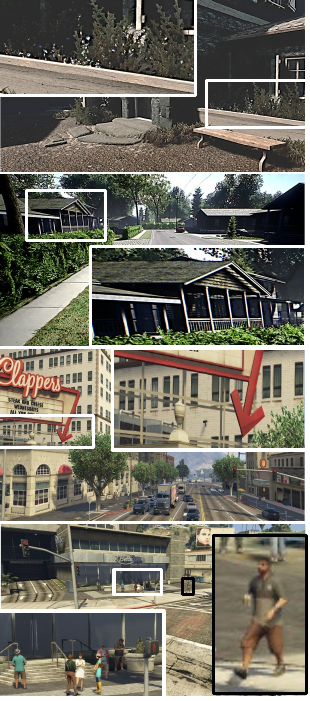} &
    \includegraphics[width=1\linewidth]{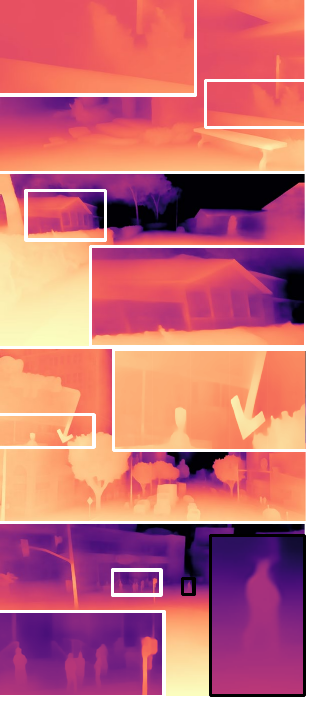} &
    \includegraphics[width=1\linewidth]{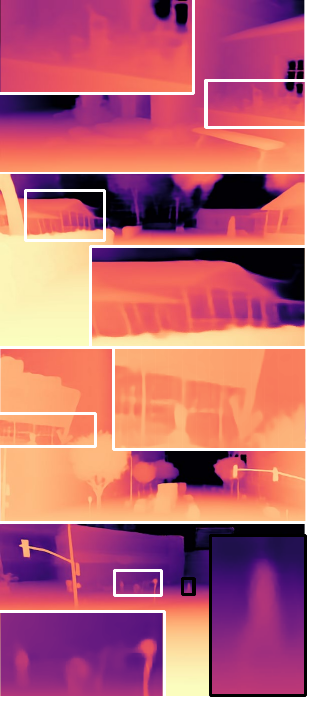} &
    \includegraphics[width=1\linewidth]{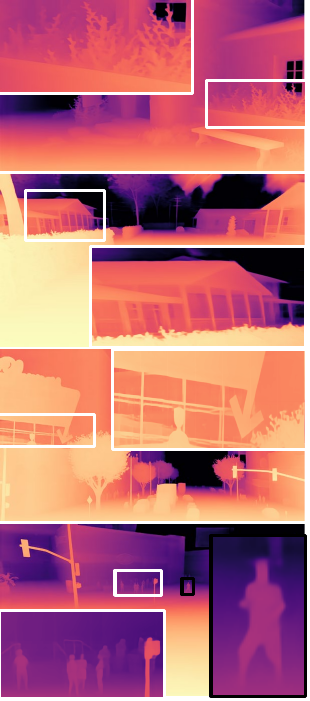} &
    \includegraphics[width=1\linewidth]{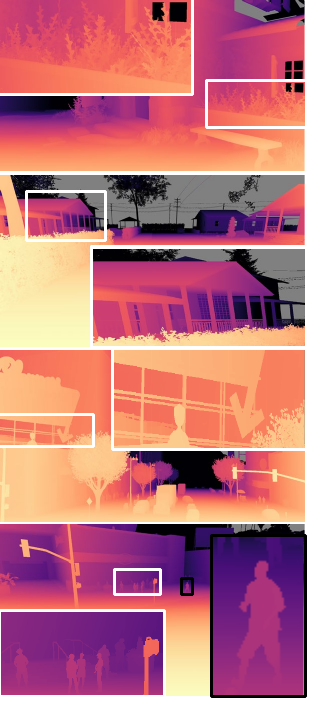} \\
    
    Input & BoostingDepth~\cite{miangoleh2021boostingdepth} & Graph-GDSR~\cite{de2022gdsr} & Ours & GT \\
    \end{tabular}
    \caption{\textbf{Qualitative Results on UnrealStereo4K and MVS-Synth.} The qualitative comparisons showcased here are derived from the UnrealStereo4K and MVS-Synth datasets, depicted in the first two and last two rows, respectively.  Zoom in to better perceive details near boundaries. Our framework outperforms counterparts~\cite{miangoleh2021boostingdepth,de2022gdsr} with much sharper edges.}
    \label{fig:u4k+gta}
\end{figure*}
Both these patches, $I_1$ and $I_2$, are independently processed through our framework, yielding image features $\mathcal{F}_g^1$ and $\mathcal{F}_g^2$, and depth predictions $\mathbf{D}_1$ and $\mathbf{D}_2$, respectively.
To enforce consistency, we impose an $L_2$ loss on the overlapping regions of both the extracted image features and the depth predictions. This loss penalizes discrepancies in the feature representations and depth predictions of the overlapping regions from the two cropped patches. The consistency-aware loss can be mathematically expressed as:
\begin{equation}
\label{eq:consistency}
\mathcal{L}_{c} = ||\mathcal{F}_g^1 - \mathcal{F}_g^2||_2 + \mu_1 ||\mathbf{D}_1 - \mathbf{D}_2||_2, ~x\in \Omega,
\end{equation}
where $c$ is short for consistency, $\mu_1$ is a hyperparameter empirically set to 0.1 to balance the loss. While the idea of constraining the depth values is quite intuitive, the good results mainly stem from constraining the features.

\textbf{Inference}: One of the distinct advantages of our method, especially when compared with BoostingDepth~\cite{miangoleh2021boostingdepth}, is its freedom from heuristic patch selection and post-processing strategies during inference. 
In our standard inference pipeline, we slice the image into $\textsc{P}=16$ non-overlapping patches, spanning its entirety. These patches are processed, and the depth maps are seamlessly stitched together. This standard pipeline is called PatchFusion$_{\textsc{P=16}}$ in Tab.~\ref{tab:overall}.

To unlock the full power of the network, we complement consistency-aware inference in the following manner.
Our model is further amplified with the inclusion of an extra 33 shifted, tidily arranged patches, as illustrated in the supplementary material, as PatchFusion$_{\textsc{P=49}}$. 
An additional improvement is to use extra randomly sampled patches, resulting in PatchFusion$_{\textsc{R}}$. Unless otherwise specified, we use $R=128$ random patches in our experiments.

During the processing of patches, the updated depth $\mathbf{D}_g$ is concatenated with the cropped image $I$ and coarse depth map $\mathbf{D}_c$, supplanting the $\mathbf{D}_f$, as the input to our guided fusion network. This dynamic updating, coupled with a running mean, engenders a local ensemble\footnote{Ensemble is combining several different predictions from different models to make the final prediction. Here, we abuse this term to represent combining different predictions from different patches but using the same model} approach, incrementally refining the depth estimations on the fly.

\subsection{Implementation Details}
\label{subsec:train&infer}

\textbf{Training}: Both networks, $\mathcal{N}_{c}$ and $\mathcal{N}_{f}$, are trained utilizing the scale-invariant loss $\mathcal{L}_{si}$~\cite{eigen2014mde,bhat2023zoedepth,li2022binsformer}. We use weights from pretraining on the NYU-v2 dataset~\cite{silberman2012nyu} as our initialization. $\mathcal{N}_{c}$ and $\mathcal{N}_{f}$ are then fine-tuned on the target dataset for 16 epochs. The training of the guided fusion network involves a combination of the scale-invariant loss $\mathcal{L}_{si}$ and our specially designed consistency loss $\mathcal{L}_{c}$ as
\begin{equation}
\label{eq:overall}
\mathcal{L} = \mathcal{L}_{si} + \mu_2 \mathcal{L}_{c},
\end{equation}
where $\mu_2$ is a hyperparameter empirically set to 0.1 to balance the loss. The fusion network $\mathcal{N}_{g}$ is trained for 12 epochs. Data augmentation for $\mathcal{N}_{f}$ and $\mathcal{N}_{g}$ includes random cropping. Beyond this, we use the default augmentation strategies adopted in the baseline depth model.

\section{Experimental Results}
\label{sec:experiments}

\subsection{Datasets and Metrics}

\indent \textbf{UnrealStereo4K}: The UnrealStereo4K dataset~\cite{tosi2021smd} offers synthetic stereo images in 4K resolution (2160$\times$3840), complete with accurate, pixel-wise ground truth. Since the dataset has some incorrectly labeled images, we use the Structural Similarity Index (SSIM)~\cite{wang2004ssim} for quality assurance, comparing the original and reconstructed left images with the given disparity maps. Entries with SSIM below 0.7 are excluded (we filter 131 out of 7860 images total). Utilizing the provided camera parameters, we convert the disparity maps to metric depth maps. Our experiments follow the prescribed dataset splits in~\cite{tosi2021smd}, and we select a patch size of 540$\times$960 by default.

\textbf{MVS-Synth}: MVS-Synth~\cite{huangDeepMVS2018} is a synthetic dataset designed for training Multi-View Stereo (MVS) algorithms. It contains 120 unique sequences, each with 100 frames depicting urban scenes from the virtual environment of Grand Theft Auto V. Each frame provides a 2K (1080$\times$1920) RGB image, along with a corresponding ground truth depth map and camera parameters. We divide the dataset into 11,160 training samples and 240 validation pairs, cropping images into 270$\times$480 patches for inputs to $\mathcal{N}_{f}$ and $\mathcal{N}_{g}$.

\textbf{Middlebury 2014}: The Middlebury 2014 dataset~\cite{scharstein2014mid} comprises a set of high-resolution stereo images (almost 4K), showcasing indoor scenes in controlled lighting settings. It includes 23 images paired with corresponding ground-truth disparity maps. Direct training of a monocular metric depth estimation model on this dataset is not advised due to the risk of overfitting. We use the dataset to test our zero-shot generalization capability, particularly from synthetic to real-world scenarios.

\textbf{Metrics}: We use the standard metric depth evaluation metrics proposed in ~\cite{eigen2014mde,piccinelli2023idisc,bhat2023zoedepth} and present details in the supplementary material. Furthermore, we introduce two additional metrics to specifically evaluate the precision at object boundaries and the consistency across patches: (1) Soft Edge Error (SEE): As per the recommendations in~\cite{tosi2021smd, chen2019over}, SEE assesses the fidelity of boundary estimation. It measures the discrepancy between predicted disparity and the ground truth within a 3×3 local patch around object edges, penalizing any over-smoothing tendencies and underlining the model's ability to capture high-fidelity depth contours. (2) Consistency Error (CE): To ascertain patch coherence, CE is calculated as the mean absolute error among patches designed with half-resolution overlaps. This metric evaluates the uniformity of the depth estimation across different patches, ensuring a seamless transition and consistency in the reconstructed depth map.


\begin{table*}[t!]
    \centering
    \scalebox{0.8}{
    \begin{tabular}{l|*{5}{C{1.4cm}}|*{5}{C{1.4cm}}}
        \toprule
        \multirow{2}{*}{Method}  & \multicolumn{5}{c|}{UnrealStereo4K} & \multicolumn{5}{c}{MVS-Synth}\\
        
        & \boldsymbol{$\delta_1 (\%)$}$\uparrow$ & \textbf{REL}$\downarrow$ & \textbf{RMS}$\downarrow$ & \textbf{SiLog}$\downarrow$ & \textbf{SEE}$\downarrow$ & \boldsymbol{$\delta_1 (\%)$}$\uparrow$ & \textbf{REL}$\downarrow$ & \textbf{RMS}$\downarrow$ & \textbf{SiLog}$\downarrow$ & \textbf{SEE}$\downarrow$  \\
        \midrule
        iDisc~\cite{piccinelli2023idisc}          & 96.940 & 0.0534 & 1.4035 & 8.5022 & 1.0697  & 93.010  & 0.0866 & 1.4386 & 15.4157 & 1.5624 \\
        BoostingDepth*~\cite{miangoleh2021boostingdepth}  & 75.483 & 0.1890 & 4.7310 & 56.3251 & 3.3204  & 71.393  & 0.2731 & 4.6859  & 85.8841 & 4.1082 \\
        BoostingDepth~\cite{miangoleh2021boostingdepth} & 98.104 & 0.0437 & 1.1233 & 6.6623 & 0.9390 & 95.409  & 0.0694 & 0.9535 & 11.5144 & 1.2694 \\
        Graph-GDSR*~\cite{de2022gdsr} & 97.757 & 0.0454 & 1.3012 & 7.6316 & 0.8734 & 94.075 &  0.0760 & 1.2735 & 14.8825 & 1.2386 \\
        Graph-GDSR~\cite{de2022gdsr} & 97.932 & 0.0438 & 1.2642 & 7.4691 & 0.8718 & 94.195 & 0.0748 & 1.2435 & 14.0723 & 1.2106 \\
        SMD-Net~\cite{tosi2021smd} & 97.774 & 0.0439 & 1.2817 & 7.3888 &  0.8828 & 93.842 & 0.0776 & 1.2563 & 14.1074 & 1.2747 \\
        \midrule
        ZoeDepth$_{\textsc{coarse}}$~\cite{bhat2023zoedepth}                 & 97.717 & 0.0455 & 1.2887 & 9.1227 & 0.9144 & 93.978 & 0.0769 & 1.2676 & 14.1236 & 1.3036 \\
        ZoeDepth$_{\textsc{fine}}$~\cite{bhat2023zoedepth}                   & 97.027 & 0.0627 & 1.2058 & 7.4483 & 0.9546 & 95.113 & 0.0715 & 0.9454 & 11.3844 & 1.1142 \\
        ZoeDepth+PF$_{\textsc{P=16}}$ & 98.419 & 0.0399 & 1.0878 & 6.2122 & \textbf{0.8382} & 95.991 & 0.0613 & 0.9213  & 10.1511 & \underline{1.0759} \\
        ZoeDepth+PF$_{\textsc{P=49}}$ & \underline{98.450} & \underline{0.0392} & \underline{1.0747} & \underline{6.1311} & \underline{0.8462} & \underline{96.069} & \underline{0.0599} & \underline{0.9050} & \underline{9.9524} & \textbf{1.0700} \\
        ZoeDepth+PF$_{\textsc{R=128}}$  & \textbf{98.469} & \textbf{0.0388} & \textbf{1.0655} & \textbf{6.0846} & 0.8488 & \textbf{96.172} & \textbf{0.0589} & \textbf{0.8944} & \textbf{9.7696} & 1.0833 \\
        \bottomrule
    \end{tabular}
    }
    \caption{\textbf{Quantitative comparison on UnrealStereo4K and MVS-Synth.} Best results are in \textbf{bold}, second best are \underline{underlined}. * indicates out of bbox inference without training on the target datasets. PF is short for PatchFusion.}
    \label{tab:overall}
\end{table*}


\begin{figure*}
\setlength\tabcolsep{1pt}
\centering
\small
    \begin{tabular}{@{}L{4.3cm}L{4.3cm}L{4.3cm}L{4.3cm}@{}}
    \multicolumn{4}{c}{\includegraphics[width=0.98\linewidth]{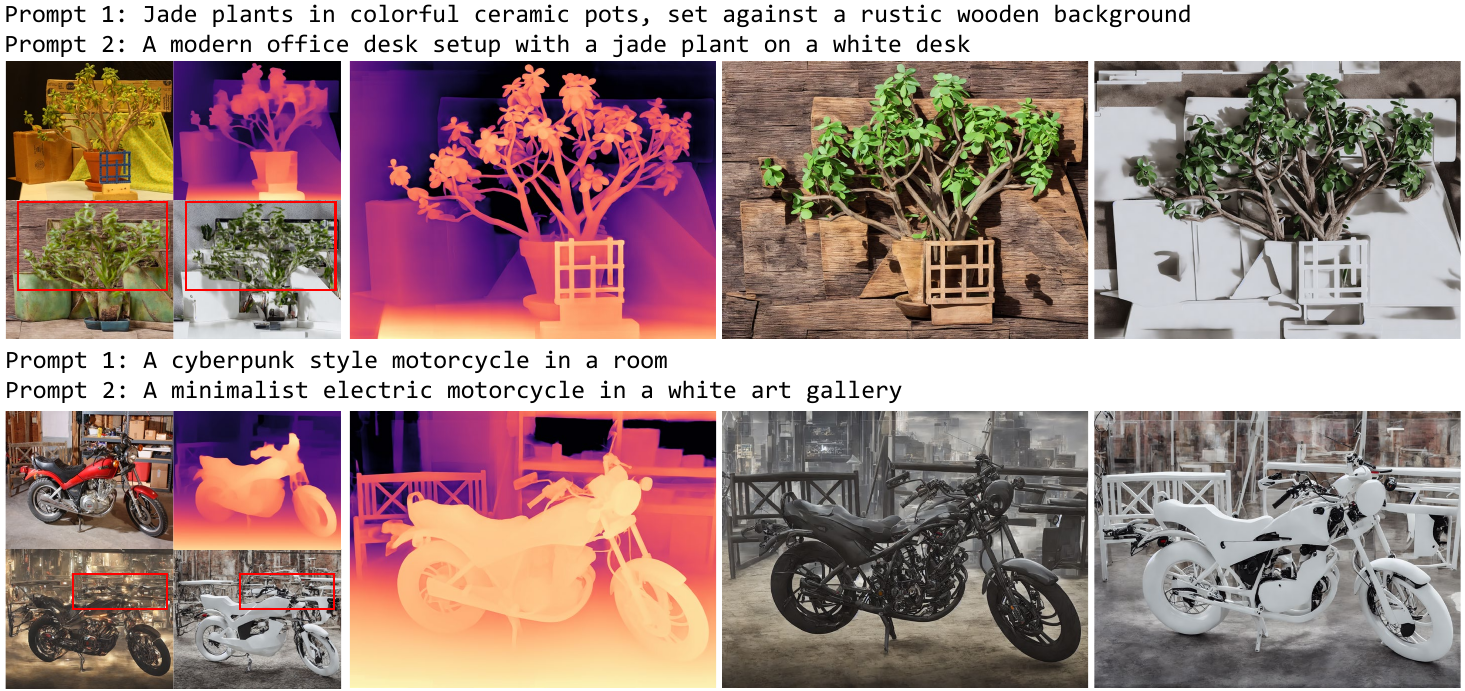}} \\
    \hspace{0.7cm} Input, Base Depth~\cite{Ranftl2022midas} & \hspace{0.9cm} Our Depth Map & \hspace{0.6cm} Prompt 1 + Our Depth & \hspace{0.5cm} Prompt 2 + Our Depth \\
    \end{tabular}
    \caption{\textbf{Qualitative Results of Depth-Guided Text-to-Image Generation.} We show two cases each with two different text prompts. Given the image, we use the default depth estimation model~\cite{Ranftl2022midas} in ControlNet~\cite{zhang2023controlnet} and our PatchFusion to predict depth maps, respectively. Along with the same prompt, the depth maps are processed by the text-to-image model in ControlNet to generate images. The depth resolution is aligned by rescaling. We show some images in a smaller size to save space. Zoom in for details like the leaves and handlebar marked by \textcolor{red}{red} boxes. The images are from Middlebury 2014~\cite{scharstein2014mid}.}
    \label{fig:control}
\end{figure*}

\begin{table}[t!]
    \centering
    \scalebox{0.68}{
    \begin{tabular}{C{0.2cm}|*{4}{C{0.95cm}}|*{3}{C{1.0cm}}|C{1.0cm}}
        \toprule
        &\multicolumn{4}{r|}{Method}  & \textbf{REL}$\downarrow$  & \textbf{SEE}$\downarrow$ & \textbf{CE}$\downarrow$ & \textbf{T}(s) $\downarrow$\\
        \midrule
        &\multicolumn{4}{r|}{Coarse Baseline} &   0.0455 & 0.9144 &  - & 0.080 \\
        &\multicolumn{4}{r|}{Fine Baseline}   &   0.0627 & 0.9546  &  0.2546 & 1.132 \\
        \midrule
        \ding{172}&\multicolumn{4}{r|}{Patch Global Scale\&Shift opt.} &   0.0544 & 1.0060 & 0.3703 & 2.256 \\
        \ding{173}&\multicolumn{4}{r|}{Pixel Deformation opt.} &   0.0437 & 0.9016 & 0.2213 & 9.729 \\
        \ding{174}&\multicolumn{4}{r|}{Poisson Image Editing opt.}        &   0.0467 & 1.0313 & 0.2499 & 31.875 \\
        \midrule
        & G & G2L & CAT-P & CAT-F \\
        \midrule
        \ding{175} & & & & & 0.0450 & 0.8932 & 0.2155 & 1.556 \\
        \midrule
        \ding{176} & \checkmark &            &            &            & 0.0423 & 0.8588 & 0.2318 & \multirow{2}{*}{1.642} \\
        \ding{177} & \checkmark &            & \checkmark &            & 0.0419 & 0.8415 & 0.1700 &                        \\
        \midrule 
        \ding{178} & \checkmark & \checkmark &            &            & 0.0414 & 0.8473 & 0.1714 & \multirow{4}{*}{2.782} \\
        \ding{179} & \checkmark & \checkmark & \checkmark &            & 0.0411 & 0.8624 & \underline{0.1215} &                        \\
        \ding{180} & \checkmark & \checkmark &            & \checkmark & 0.0403 & 0.8451 & 0.1624 &                        \\
        \ding{181} & \checkmark & \checkmark & \checkmark & \checkmark & 0.0399 & \textbf{0.8382} & 0.1441 &                        \\
        \midrule
         & \multicolumn{4}{r|}{PatchFusion$_{\textsc{P}=49}$ (w/o CAI)} & \underline{0.0398} & \underline{0.8397} & 0.1441 &      \multirow{2}{*}{7.818}  \\
        \ding{72} & \multicolumn{4}{r|}{PatchFusion$_{\textsc{P}=49}$ (full method)} & \textbf{0.0392} & 0.8462 & \textbf{0.0464} &                        \\
        \bottomrule
    \end{tabular} 
    }
    \caption{Ablation study on UnrealStereo4K dataset. G denotes the feature guidance. CAT-P and CAT-F denote adopting consistency-aware training on predictions and features, respectively. Best results are in \textbf{bold}, second best are \underline{underlined}. Time: average inference time for one image.}
    \label{tab:abl}
\end{table}





\subsection{Main Results}
Our framework can enhance any metric depth estimation model. For our experiments, we select ZoeDepth~\cite{bhat2023zoedepth} as our baseline due to its SOTA performance. We compare our framework against various other methods, including (1) traditional monocular metric depth estimation methods such as ZoeDepth~\cite{bhat2023zoedepth} and iDisc~\cite{piccinelli2023idisc}, (2) the guided depth super-resolution approach Graph-GDSR~\cite{de2022gdsr}, (3) SMD-Net~\cite{tosi2021smd} which utilizes an implicit function head for enhanced depth estimation, and (4) the tile-based technique BoostingDepth~\cite{miangoleh2021boostingdepth}. To ensure fair comparisons, we standardize the depth models across these strategies to ZoeDepth or we fine-tune the models using their provided NYU-v2~\cite{silberman2012nyu} pretrained versions on both the UnrealStereo4K~\cite{tosi2021smd} and MVS-Synth~\cite{huangDeepMVS2018} datasets, setting this as our default experimental protocol.

\textbf{UnrealStereo4K}: The performance comparison on the UnrealStereo4K test set is presented on the left of Tab.~\ref{tab:overall}. While all compared methods enhance prediction quality to some degree, our PatchFusion framework significantly surpasses the baseline ZoeDepth model by 17.3\% in RMSE and 14.7\% in REL. This indicates a considerable performance gain afforded by our approach. Our results also showcase the lowest Soft Edge Error (SEE), evidencing superior boundary delineation. Qualitative comparisons in Fig.~\ref{fig:u4k+gta} further underscore the high-quality depth maps generated by our framework. We capture intricate details, especially in boundary areas. More cases are shown in our supplementary material.

\textbf{MVS-Synth}: The right of Tab.~\ref{tab:overall} provides the quantitative comparison results for experiments on the MVS-Synth dataset~\cite{huangDeepMVS2018}. Our framework outperforms all alternative approaches in every evaluated metric. Notably, our technique boosts the RMS score by approximately 29.4\% and reduces the scale-invariant logarithmic error by 30.8\%. We present qualitative results in Fig.~\ref{fig:u4k+gta} and the supplementary material.

{\textbf{Middlebury 2014}: We first evaluate the zero-shot transfer capability of our framework. The accompanying supplementary material provides both quantitative and qualitative comparisons to illustrate the efficacy of our approach. Building on this foundation, we explored our framework's application to the field of text-to-image generation. We substituted the conventional depth estimation model~\cite{Ranftl2022midas} employed by ControlNet~\cite{zhang2023controlnet} with our PatchFusion. To ensure compatibility with the depth range required by ControlNet, we normalized the metric depth output of PatchFusion to the interval $[0, 256]$ and then applied the transformation $256-\hat{d}$ to invert it. The results are illustrated in Fig.~\ref{fig:control}. Armed with the high-fidelity depth maps generated by PatchFusion, which capture intricate details, the text-to-image generation model yields coherent structures, textures, and details of significantly higher quality than the baseline.

\subsection{Ablation Studies and Discussion}
\label{sec:subsec:ablation}

In this section, we ablate and discuss the contributions of individual components within our framework. Unless stated otherwise, we utilize the UnrealStereo4K dataset and the PatchFusion variant with $P=16$ patches for clarity and ease of comparison. Inference time benchmarks are performed on a single NVIDIA A100 GPU.

\textbf{Comparison with other Blending Strategies:}
The primary role of the fusion network in our framework is to act as an adaptive blending module, integrating the fine detail from $\mathbf{D}_f$ with the scale information retained in $\mathbf{D}_c$. We benchmark our network against other blending strategies prevalent in various tile-based depth estimation methods~\cite{miangoleh2021boostingdepth, rey2022360monodepthtile, hedman2018instant}. As illustrated in Tab.~\ref{tab:abl} (\ding{172}, \ding{173}, \ding{174}), our approach yields superior results in terms of estimation accuracy and satisfactory computational efficiency. We highlight that traditional blending strategies often grapple with the challenge of specifying an optimal handcrafted target for optimization, which can result in suboptimal outcomes. Our network circumvents this issue by learning to blend the input patches effectively through a data-driven process. This is evidenced by the substantial gains in performance metrics, with an 8.7\% improvement in REL and a 34.8\% reduction in CE during inference.

\textbf{Guided Fusion Network with G2L Module:}
Our initial focus is to evaluate the impact of the Guided Fusion Network equipped with the G2L (Global to Local) Module. We incrementally introduce the guidance feature and the G2L module to a baseline encoder-decoder structure, similar to the MergeNet in BoostingDepth~\cite{miangoleh2021boostingdepth}, and observe the changes in performance. As indicated in Tab.~\ref{tab:abl}, the standalone model without the guidance feature and the G2L module (\ding{175}) yields the least impressive results. With RMSE and SEE metrics closely aligning with the coarse baseline, and only a marginal improvement in the consistency error, it becomes evident that this variant struggles to integrate the finer details from $\mathbf{D}_f$ into the coarse depth map $\mathbf{D}_c$. This underlines the crucial role our framework plays in surpassing the capabilities of BoostingDepth~\cite{miangoleh2021boostingdepth}.

Introducing the guidance feature (\ding{176}) marks an uptick in performance with improvements of 6.0\% in REL and 3.8\% in SEE, confirming the feature's role in enhancing training and the assimilation of fine details. The integration of the G2L module (\ding{178}) leads to a 26.0\% reduction in consistency error, achieved without imposing explicit constraints. The concomitant reduction in 2.1\% REL underscores the G2L module's effectiveness in harnessing global information, thus improving both accuracy and consistency.

\textbf{Consistency-Aware Training and Inference:}
Integrating the consistency-aware training (CAT) loss, results in enhanced REL and CE metrics for models, both with and without the G2L module (\ding{177} and \ding{179}, respectively). Notably, the introduction of the G2L module and consistency loss leads to a marginal reduction in SEE. This suggests that the model may be excessively penalized, potentially neglecting some fine details in the process. When the consistency loss is applied to the intermediate features (\ding{180}), we observe a more pronounced improvement in REL, while SEE is preserved, indicating a balanced detail capture. Nevertheless, this approach yields a diminished CE, suggesting a slight trade-off. Upon concurrently enforcing consistency constraints on both predictions and intermediate features, we note a further refinement in both REL and SEE metrics, along with an acceptable level of CE (\ding{181}). This finding underscores the balanced improvements in standard metrics and patch-wise consistency, which is pivotal for a tile-based framework. Collectively, these results affirm the CAT's role in bolstering high-quality depth estimation.
Finally, with the use of additional patches and our consistency-aware inference (CAI), we obtain another significant boost in REL and CE (\ding{72}) due to the local ensemble, but at the expense of SEE and time complexity.

\subsection{Limitations and Future Work}

One limitation is the computational efficiency of our framework. While incorporating an increasing number of randomly selected patches does improve depth prediction, it also results in a proportional increase in processing time. We therefore suggest an efficient patch selection strategy as an avenue for future work. Another limitation is the lack of high-resolution real-world training data. We can observe this gap in our results from synthetic to real transfer. While the results showcase sharp edges, the scale of the results could be improved. Also, we noticed that UnrealStereo4K~\cite{tosi2021smd} does not contain enough images with large homogeneous foreground objects. We believe that the collection of large real-world high-resolution depth datasets will be a great contribution, as depth estimation, in general, seems to be a highly valuable pre-training task~\cite{goldblum2023battle}, and therefore recommend this as future work.

\section{Conclusions}
\label{sec:conclusion}

We presented \textbf{PatchFusion}, an end-to-end tile-based framework tailored to high-resolution monocular metric depth estimation. We introduced a novel tile-based network architecture together with a consistency-aware training and inference strategy. This combination yields a framework that only relies on forward passes through networks and obviates the need for additional pre-processing and post-processing. Our proposed framework decisively improves upon the baseline model for UnrealStereo4K (17.3\% in RMSE) and MVS-Synth (29.4\% in RMSE), while demonstrating satisfactory performance in zero-shot transfer.

\appendix

        
            
            

\begin{figure*}[h]
\setlength\tabcolsep{1pt}
\centering
\small
    \begin{tabular}{@{}*{5}{C{3.4cm}}@{}}
    \includegraphics[width=1\linewidth]{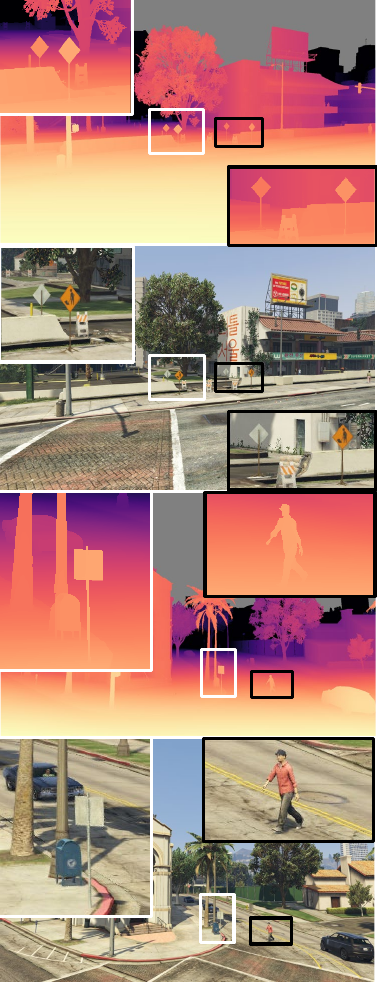} &
    \includegraphics[width=1\linewidth]{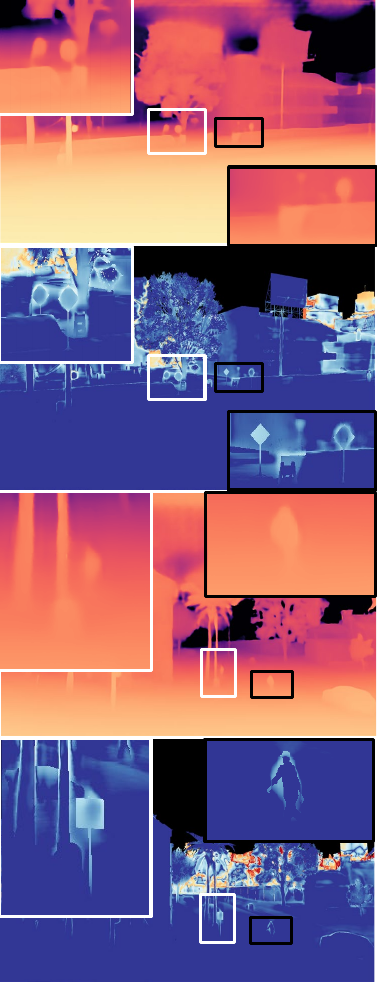} &
    \includegraphics[width=1\linewidth]{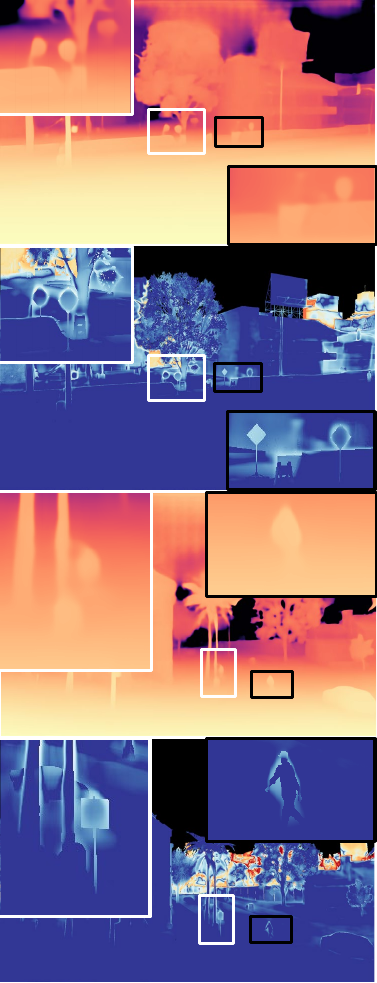} &
    \includegraphics[width=1\linewidth]{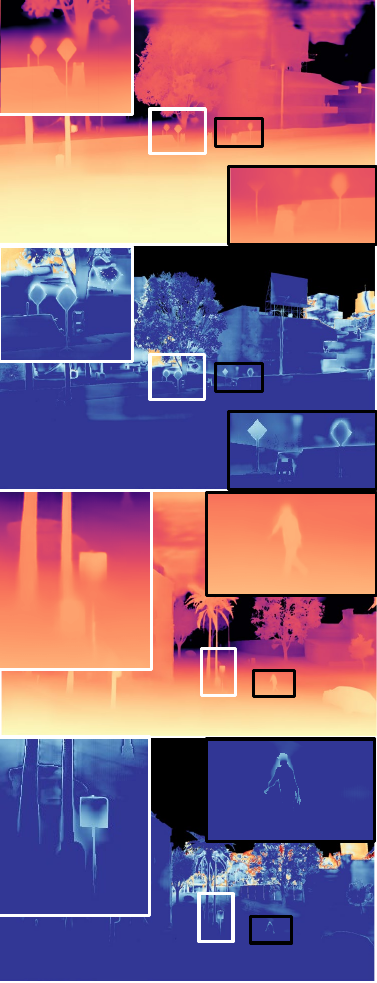} &
    \includegraphics[width=1\linewidth]{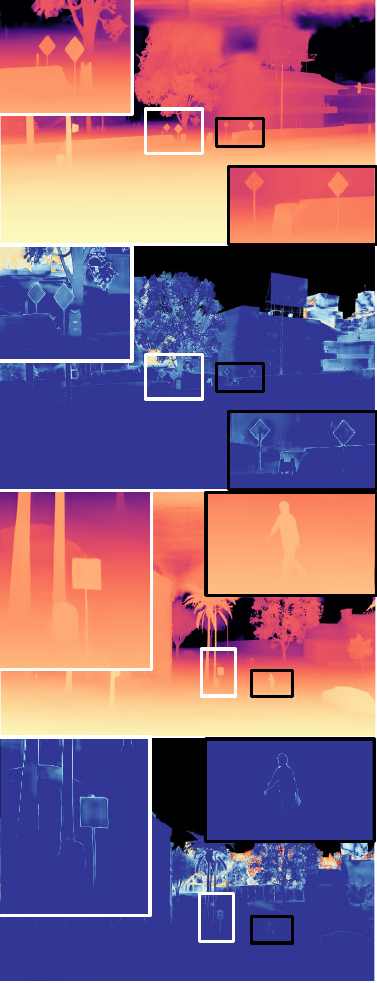} \\
    
    GT, Input & Graph-DGSR~\cite{de2022gdsr} & SMD-Net~\cite{tosi2021smd} & BoostingDepth~\cite{miangoleh2021boostingdepth} & Ours \\
    \end{tabular}
    \caption{\textbf{Qualitative Results on MVS-Synth.} The first row shows the predicted depth maps while the second row depicts the corresponding error maps. Zoom in to better perceive details near boundaries. As we claim in Sec.1, due to the low-resolution nature of base models, DGSR can propagate errors, and the implicit function still strips away crucial high-frequency details during downsampling. Compared with the SOTA tile-based framework BoostingDepth~\cite{miangoleh2021boostingdepth}, our framework achieves better estimation quality at boundaries and shows better scale consistency.}
    \label{fig:gta}
\end{figure*}

\begin{figure*}
\setlength\tabcolsep{1pt}
\centering
\small
    \begin{tabular}{@{}*{4}{C{4.2cm}}@{}}
    \includegraphics[width=1\linewidth]{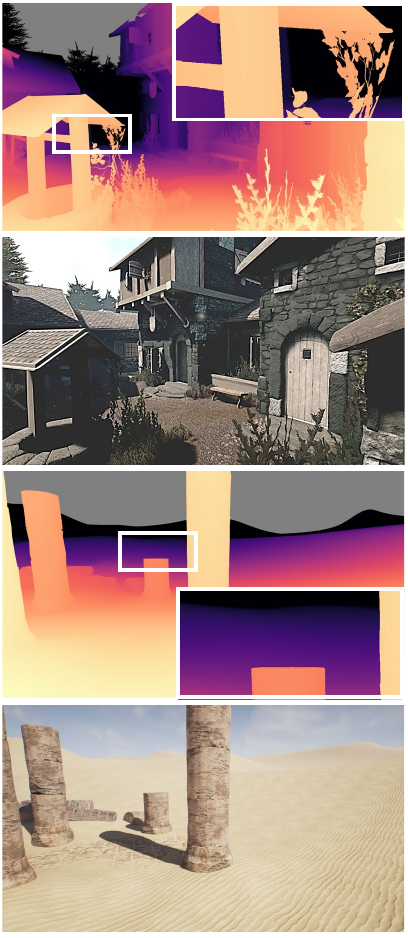} &
    \includegraphics[width=1\linewidth]{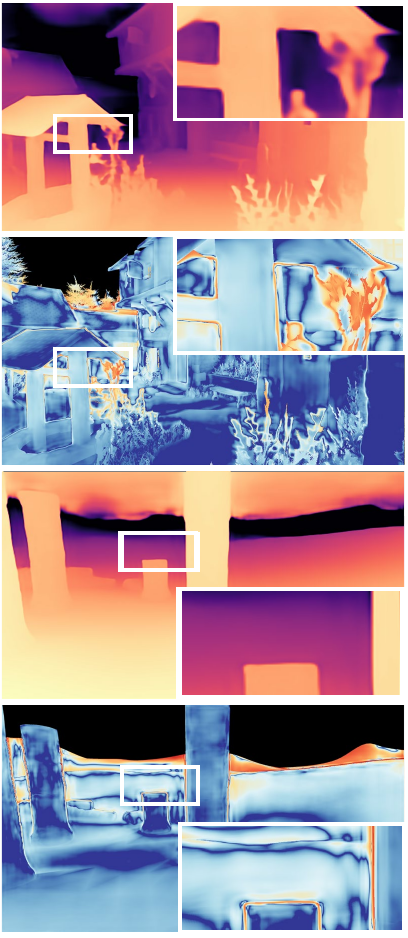} &
    \includegraphics[width=1\linewidth]{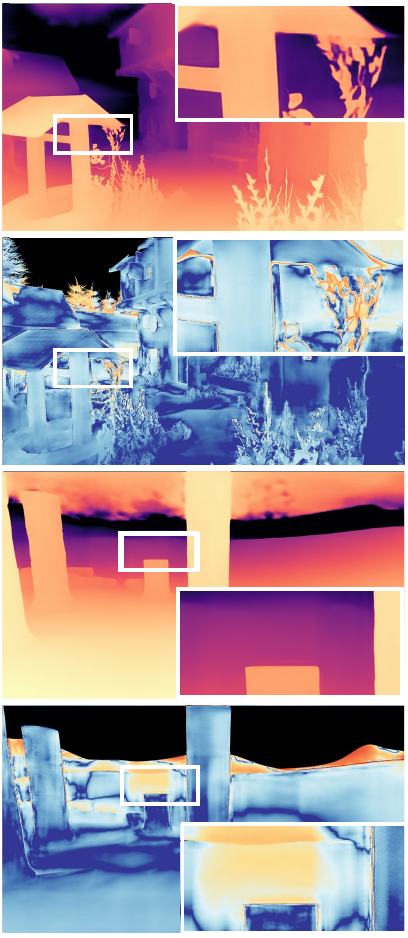} &
    \includegraphics[width=1\linewidth]{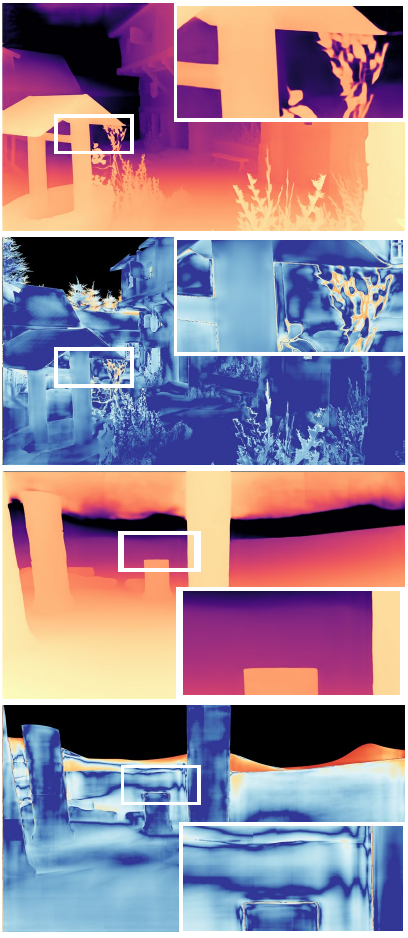} \\
    
    GT, Input & ZeoDepth$_{\textsc{COARSE}}$ & BoostingDepth~\cite{miangoleh2021boostingdepth} & Ours \\
    \end{tabular}
    \caption{\textbf{Qualitative Results on UnrealStereo4K.} The first row shows the predicted depth maps while the second row depicts the corresponding error maps. Zoom in to better perceive details near boundaries. Our framework outperforms BoostingDepth~\cite{miangoleh2021boostingdepth} not only in details but also scale-consistency.}
    \label{fig:u4k}
\end{figure*}

\section{Evaluation Metrics}
We follow the standard evaluation protocol proposed in previous monocular metric depth estimation works to evaluate the effectiveness of our framework. We utilize the root mean squared error (RMSE) $=|\frac{1}{M}\sum_{i=1}^M|d_i-\hat{d}_i|^2|^\frac{1}{2}$, mean absolute relative error (AbsRel) $=\frac{1}{M}\sum_{i=1}^M|d_i-\hat{d}_i|/d_i$, scale-invariant logarithmic error (SILog) $=|\frac{1}{M}\sum_{i=1}^Me^2-|\frac{1}{M}\sum_{i=1}^Me|^2|^{1/2} \times 100$ where $e=\log{\hat{d}_i} - \log{d_i}$, and the accuracy under the threshold ($\delta_i < 1.25^i, i = 1$), where $d_i$ and $\hat{d}_i$ refer to ground truth and predicted depth at pixel $i$, respectively, and $M$ is the total number of pixels in the image. Since UnrealStereo4K~\cite{tosi2021smd} contains both indoor and outdoor images, we cap the evaluation depth at the range of $(1e^{-3}\text{m}, 80\text{m})$. Final model outputs are evaluated at ground truth resolution.

\section{More Qualitative Results}
We present qualitative comparisons with all three types of strategies for depth super-resolution in Fig.~\ref{fig:gta}, \ref{fig:u4k}: (1) Depth-Guided Super-Resolution Graph-GDSR~\cite{de2022gdsr}, (2) Baseline depth model ZoeDepth~\cite{bhat2023zoedepth} with Implicit Function~\cite{tosi2021smd}, and (3) Tile-Based Framework BoostingDepth~\cite{miangoleh2021boostingdepth} finetuning on the target dataset. Due to the low-resolution nature of base models, (1) DGSR propagates errors, and the implicit function (2) still strips away crucial high-frequency details during input downsampling. Compared with (3) the SOTA tile-based framework BoostingDepth~\cite{miangoleh2021boostingdepth}, our framework achieves better estimation quality at boundaries and shows better scale consistency.

\begin{figure*}
\setlength\tabcolsep{1pt}
\centering
\small
    \begin{tabular}{@{}*{4}{C{3.6cm}}@{}}
    \includegraphics[width=1\linewidth]{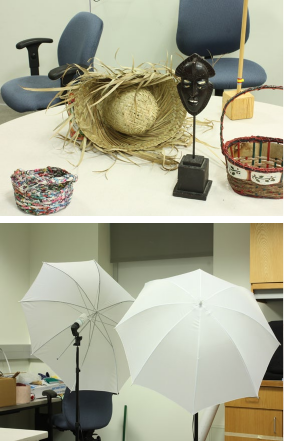} &
    \includegraphics[width=1\linewidth]{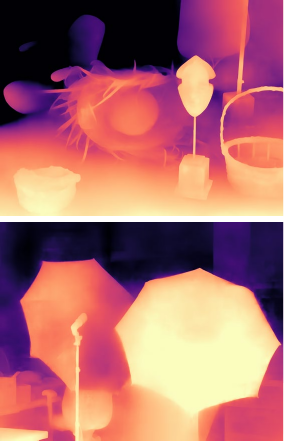} &
    \includegraphics[width=1\linewidth]{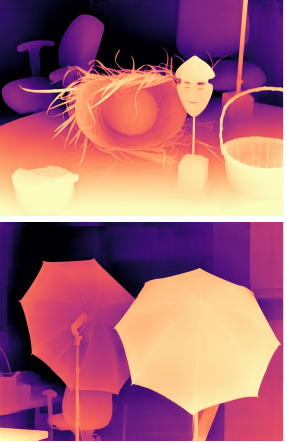} &
    \includegraphics[width=1\linewidth]{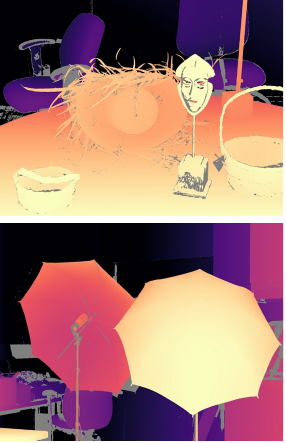} \\
    
    Input & BoostingDepth~\cite{miangoleh2021boostingdepth} & Ours & GT \\
    \end{tabular}
    \caption{\textbf{Qualitative Comparison on Middlebury 2014.} Zoom in to observe more details. We use BoostingDepth~\cite{miangoleh2021boostingdepth} out of the box and align the prediction of BoostingDepth with ground-truth scales with post-process optimization for visualization. Note that BoostingDepth is trained on this dataset, while we adopt PatchFusion in a synthetic-to-real zero-shot transfer manner.}
    \label{fig:zero-shot}
\end{figure*}

\section{Zero-Shot Transfer on Middlebury 2014}

\begin{table}[t!]
    \centering
    \scalebox{0.75}{
    \begin{tabular}{L{4.7cm}|*{2}{C{1.5cm}}}
        \toprule
        Method  & \textbf{RMS}$\downarrow$ & \textbf{SEE}$\downarrow$  \\
        \midrule
        ZoeDepth$_{\textsc{COARSE}}$  & 1.0777 & 0.8326 \\
        \midrule
        ZoeDepth + PatchFusion$_{\textsc{P=16}}$ & 1.0743 & 0.8284  \\
        ZoeDepth + PatchFusion$_{\textsc{P=40}}$ & 1.0678 & 0.8219  \\
        ZoeDepth + PatchFusion$_{\textsc{r=128}}$ & 1.0620 & 0.8195  \\
        ZoeDepth + PatchFusion$_{\textsc{r=256}}$ & \underline{1.0580} & \underline{0.8194}  \\
        ZoeDepth + PatchFusion$_{\textsc{r=1024}}$ & \textbf{1.0536} & \textbf{0.8178} \\
        \bottomrule
    \end{tabular} 
    }
    \caption{Results of models trained on the UnrealStereo4K dataset and tested on the Middleburry 2014 dataset without fine-tuning. Best results are in \textbf{bold}, second best are \underline{underlined}.}
    \label{tab:zeroshot}
\end{table}

In this set of experiments, we delve into the zero-shot transfer capability of our framework, which is an essential quality for deployment in diverse real-world scenarios. The outcomes, as depicted in Tab.~\ref{tab:zeroshot}, confirm that each configuration of our framework successfully elevates the zero-shot transfer potential beyond the baseline model. The comparisons also substantiate our claim from Sec.~3: incorporating an incremental number of randomly selected patches within our framework would systematically refine the depth estimations. Visual results are provided in Fig.~\ref{fig:zero-shot}. Remarkably, our method achieves enhanced results on the Middlebury 2014 dataset~\cite{scharstein2014mid}, delineating clearer object boundaries, despite the fact that it has not been trained on it. This is in contrast to the BoostingDepth~\cite{miangoleh2021boostingdepth}, which includes training on this specific dataset.

\section{Inference Setup}

As shown in Fig.~\ref{fig:sup-infer}, we slice the image into $\textsc{P}=16$ non-overlapping patches, spanning its entirety. These patches are processed, and the depth maps are seamlessly stitched together. This standard pipeline is called PatchFusion$_{\textsc{P=16}}$. To unlock the full power of the network, our model is further amplified with the inclusion of an extra 33 shifted, tidily arranged patches, as PatchFusion$_{\textsc{P=49}}$. An additional improvement is to use extra randomly sampled patches, resulting in PatchFusion$_{\textsc{R}}$.

\begin{figure}
    \centering
    \includegraphics[width=0.95\linewidth]{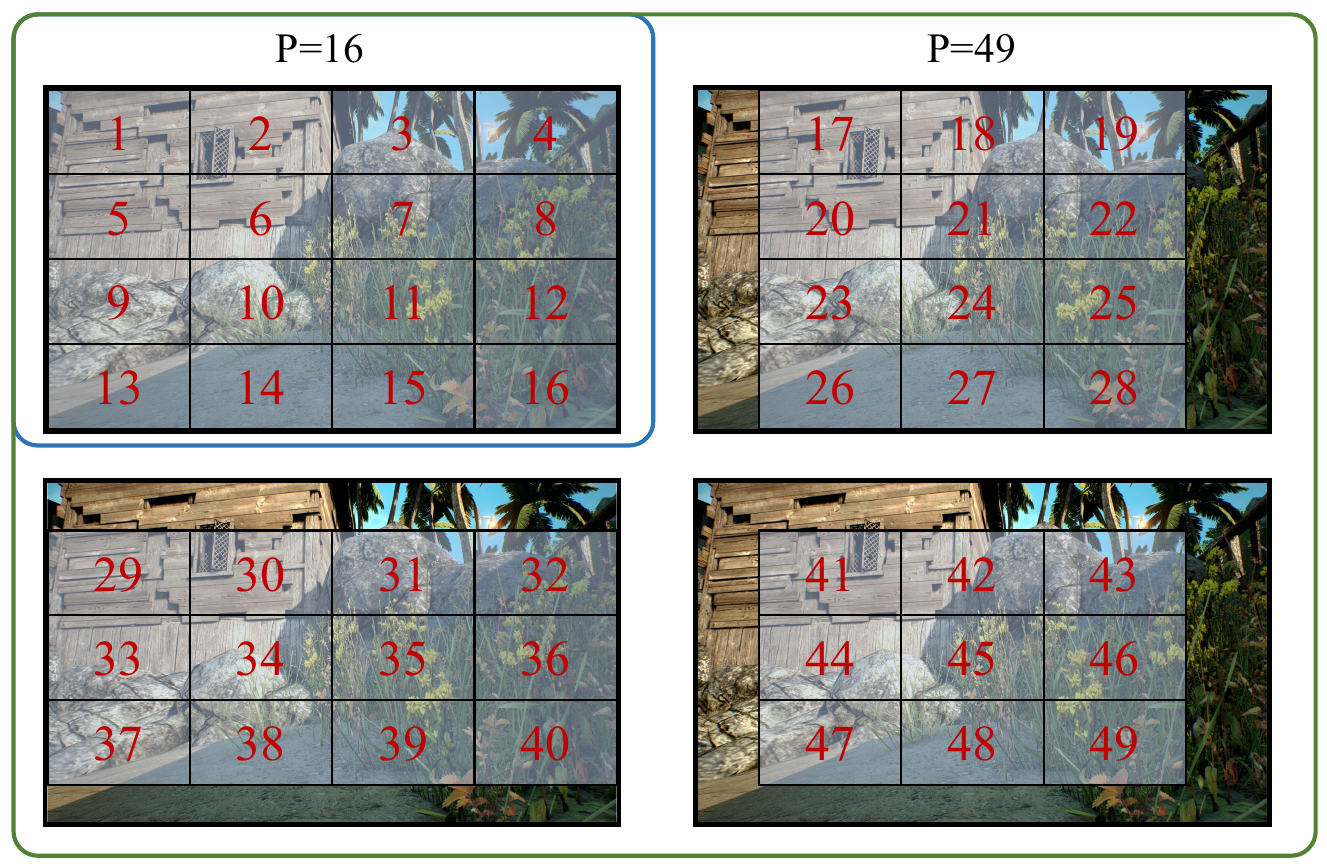}
    \caption{\textbf{Tile Placement.} We slice the image into $\textsc{P}=16$ non-overlapping patches, leading to PatchFusion$_{\textsc{P}=16}$. Then, we introduce an extra 33 shifted, tidily arranged patches, corresponding to PatchFusion$_{\textsc{P}=49}$. In PatchFusion$_{\textsc{R}=128}$, the extra 128 patches are then randomly placed on the image.}
    \label{fig:sup-infer}
\end{figure}

\section{Qualitative Comparison of CAI}

In our study, we introduce a Consistency-Aware Inference (CAI) technique aimed at improving the consistency of patch-wise depth estimations during the inference phase. The significance of this approach is highlighted in Fig.~\ref{fig:sup-cai}. Although our Guided Fusion Network, supplemented by the Global-to-Local module and Consistency-Aware Training, facilitates scale-aware and consistency-aware predictions, discontinuities, particularly around prediction edges, remain a challenge. This problem persists even when increasing the number of randomly selected patches, as newer patch estimations tend to overwrite the earlier ones.

To address this issue, the CAI strategy updates predictions in a running average manner. It calculates the prediction for each pixel as an average of all previous predictions from overlapping patches. This method not only ensures robustness but also enhances the continuity of predictions across different patches, resulting in a more seamless and integrated depth estimation.

\begin{figure}
    \centering
    \includegraphics[width=0.95\linewidth]{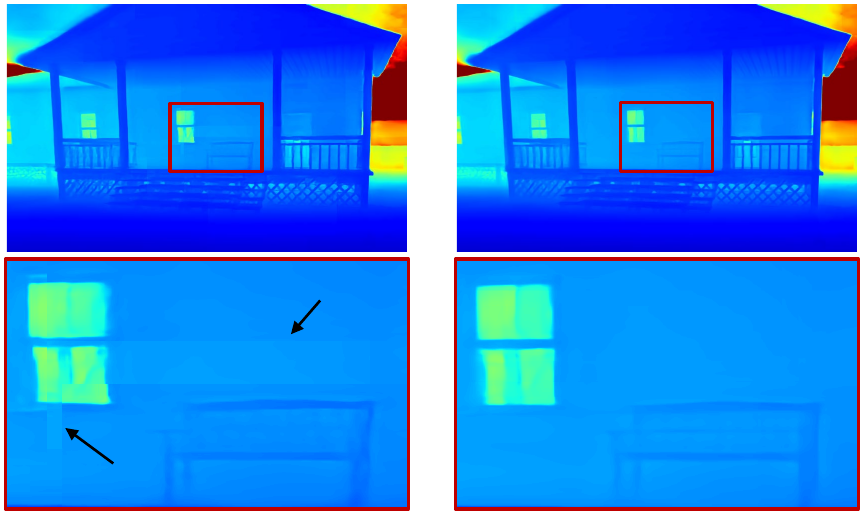}
    \caption{\textbf{Qualitative Comparison of Consistency-Aware Inference (CAI).} \textbf{Left:} PatchFusion$_{\textsc{R=128}}$ without CAI. \textbf{Right:} PatchFusion$_{\textsc{R=128}}$ with CAI. Without CAI, newer patch estimations tend to overwrite the earlier ones, leading to inevitable artifacts on patch boundaries. Our CAI successfully alleviate this issue by the running average strategy.}
    \label{fig:sup-cai}
\end{figure}

{
    \small
    \bibliographystyle{ieeenat_fullname}
    \bibliography{reference}
}

\end{document}